\documentclass[letterpaper]{article} 
\usepackage{aaai2027}
\usepackage[hyphens]{url}  
\usepackage{graphicx} 
\usepackage{natbib}  
\usepackage{caption} 
\usepackage{algorithm}
\usepackage{algorithmic}

\usepackage{newfloat}
\usepackage{listings}
\DeclareCaptionStyle{ruled}{labelfont=normalfont,labelsep=colon,strut=off} 
\floatstyle{ruled}
\newfloat{listing}{tb}{lst}{}
\floatname{listing}{Listing}

\usepackage{booktabs}

\usepackage{enumitem}
\usepackage{amssymb}
\usepackage{amsmath}
\usepackage{multirow}
\usepackage{xcolor}
\newcommand{\ours}{\textsc{GWM-VLA}}

\title{GWM-VLA: Geometry-Aware Latent World Modeling for\\Vision-Language-Action Learning}
\author{Yanping Zhao, Hang Yu, Yiwei Wang, Chen Ye, Siyu Tian, Di Zhang, Qingjun Wang, Qian Chen, Junqiao Zhao, Chen Ye, Guang Chen}
\affiliations{Tongji University}

\begin{document}

\maketitle

\begin{abstract}

Vision-Language-Action (VLA) models achieve strong robotic manipulation performance but often degrade under visual and environmental shifts. 
Latent world modeling offers a promising approach to improving robustness, yet existing methods commonly encode camera views independently and predict holistic scene dynamics without explicitly modeling their geometric relationships. 
We propose GWM-VLA, a geometry-aware latent world modeling framework for VLA learning. 
GWM-VLA combines geometry-aware multi-view state encoding, global context-conditioned target-view prediction, and shared latent-action representations grounded by robot-action supervision.
Specifically, VGGT-$\Omega$ jointly aggregates multi-view observations at each timestep to construct geometry-aware multi-view states.
The latent world model predicts the next-step patch tokens of a selected target view using patch and register tokens obtained after multi-view aggregation, thereby retaining multi-view geometric information without predicting the complete multi-view state.
We use the wrist view as the target in our experiments, placing greater emphasis on end-effector motion and local gripper-object interactions.
Finally, the shared latent-action representations condition both the latent world model and the flow-matching action head, allowing latent-prediction supervision and ground-truth robot-action supervision to jointly shape the same latent-action representations.
Experiments across both simulation and real-world environments demonstrate the effectiveness and robustness of GWM-VLA.

\end{abstract}

\section{Introduction}
\label{sec:intro}

Vision-Language-Action (VLA) models \cite{brohan2022rt,zitkovich2023rt} transfer semantic knowledge from large vision-language models to continuous robot control and have achieved strong manipulation performance across diverse tasks. However, even high-performing VLA policies can degrade substantially under controlled changes in camera viewpoint, illumination, background texture, object layout, and sensor noise \cite{fei2025libero}. This sensitivity to visual and environmental changes suggests that some policies may rely on appearance-specific correlations rather than capturing how the scene changes in response to robot actions.

Learning environment dynamics through world modeling can improve VLA
robustness by encouraging policies to capture action-induced state
changes rather than relying mainly on visual appearance.
Existing methods incorporate explicit world models in different ways.
World-action models such as DreamZero \cite{ye2026world} integrate
future visual prediction and action generation within a unified policy,
whereas $\pi_{0.7}$ \cite{pi07} uses a separate world model to generate
visual subgoals for the VLA.
However, these explicit world models require generating future visual observations.
Their prediction objectives must therefore model textures, lighting,
backgrounds, and other appearance details in addition to
control-relevant state changes.

Latent world modeling offers a more compact alternative by predicting future representations rather than generating future visual observations, reducing the need to model fine-grained appearance details. 
VLA-JEPA \cite{sun2026vla} follows this approach by predicting future visual latents conditioned on learned latent actions. 
This objective allows the world model to focus on latent state transitions without explicitly generating future images or videos.
However, VLA-JEPA uses a pretrained V-JEPA2 encoder \cite{assran2025v} to encode each camera view independently and combines the resulting features through concatenation.
Consequently, cross-view geometric relationships, including visual correspondences and camera-relative scene structure, remain implicit.

Recent feed-forward geometric foundation models \cite{wang2025vggt, wang2026vggt, leroy2024grounding, wang2024dust3r} make geometry-aware multi-view modeling practical. In particular, VGGT \cite{wang2025vggt} and VGGT-$\Omega$ \cite{wang2026vggt} jointly aggregate multiple views and encode cross-view geometric structure, making them attractive visual backbones for spatially grounded robot policies. However, how geometry-aware multi-view representations can serve as predictive states for latent world modeling remains underexplored.

\begin{figure}[t]
\centering
\includegraphics[width=0.95\linewidth]{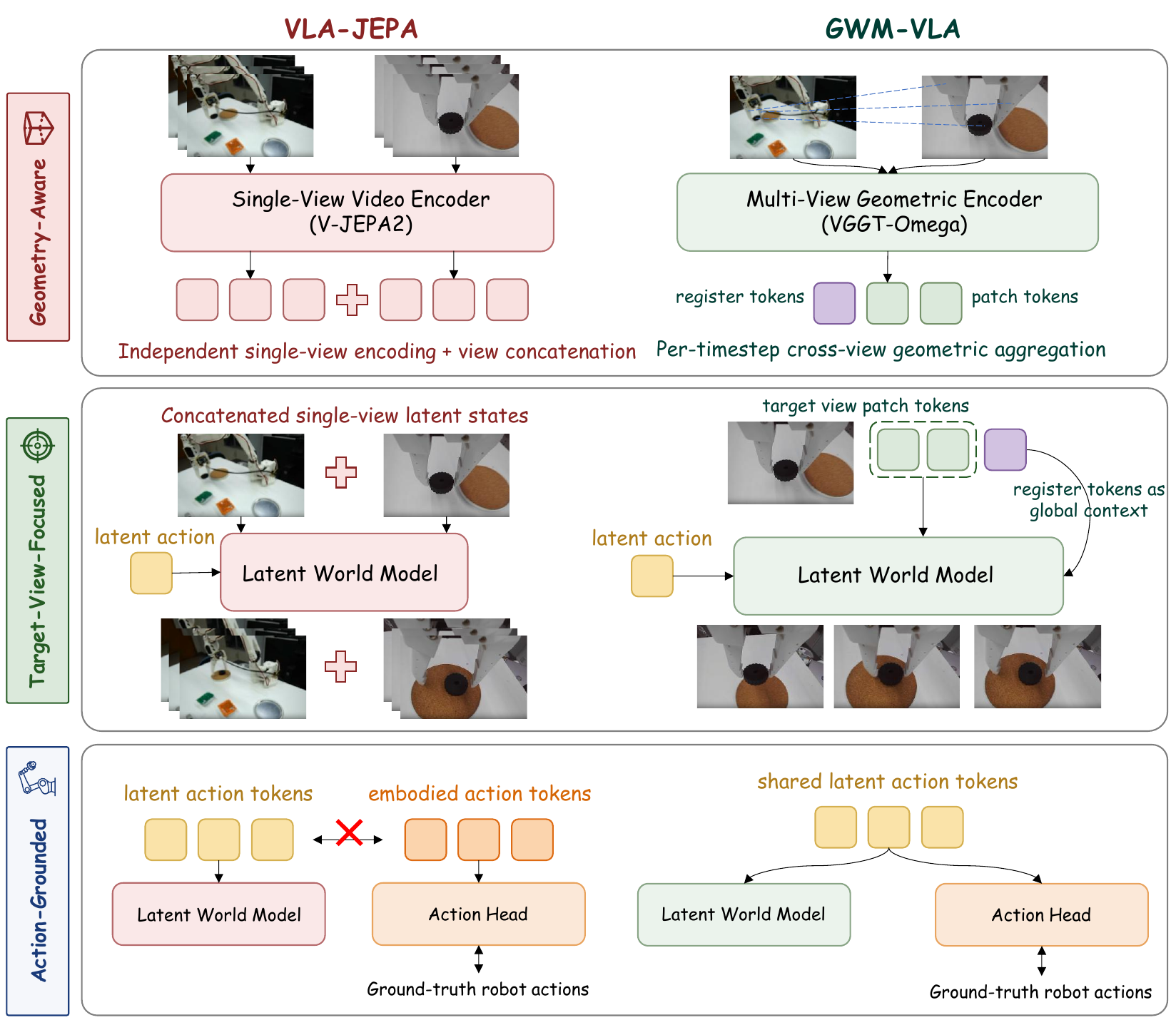}
\caption{Mechanism-level comparison between VLA-JEPA and GWM-VLA. GWM-VLA combines geometry-aware multi-view state encoding, global context-conditioned target-view prediction, and shared latent-action conditioning. }
\label{fig:motivation}
\end{figure}

To this end, we propose \textbf{GWM-VLA}, a Geometry-Aware Latent World Modeling framework for VLA learning, as illustrated in Figure~\ref{fig:motivation}.
GWM-VLA introduces three key designs.
First, VGGT-$\Omega$ aggregates multi-view observations at each timestep to construct geometry-aware multi-view states.
Second, rather than predicting the complete multi-view state, the latent world model predicts the next-step patch tokens of a selected target view, while using the corresponding target-view register tokens as global geometric context. 
This design restricts prediction to the selected view while still leveraging geometric information obtained through multi-view aggregation.
We use the wrist view in our experiments, placing greater emphasis on end-effector motion and local gripper-object interactions.
Third, the shared latent-action representations condition both the latent world model and the flow-matching action head. 
Consequently, the latent prediction and robot-action objectives jointly shape the shared latent-action representations, coupling predictive learning with continuous robot control.

We evaluate GWM-VLA in both simulation and real-world settings.
On LIBERO, GWM-VLA achieves a $97.1\%$ average success rate, matching the best reported performance while using only robot demonstrations for pretraining.
On the more challenging LIBERO-Plus robustness benchmark, GWM-VLA reaches a $76.9\%$ average success rate, outperforming the robot-only VLA-JEPA baseline by 14.0 percentage points and achieving state-of-the-art average performance.
Real-world experiments on the SO-101 robot further demonstrate that GWM-VLA achieves the highest average success rate under limited demonstrations.

Our main contributions are summarized as follows:

\begin{enumerate}[label={\bf {{$\bullet$}}}]
    \item We introduce geometry-aware latent world modeling, which uses jointly aggregated multi-view geometric representations as predictive states and models their action-conditioned temporal transitions.

    \item We propose GWM-VLA, a VLA framework that integrates geometry-aware latent world modeling with flow-matching action generation through shared latent-action representations.
    
    \item We demonstrate the effectiveness of GWM-VLA for accurate and robust robot control through simulation and real-world experiments.
\end{enumerate}

\section{Related Work}
\label{sec:related_work}

\subsection{Geometry-Aware Representations for VLA}

Geometry-aware representations are important for VLA policies because robotic manipulation requires reasoning about object geometry, spatial relationships, camera viewpoints, and gripper-object interactions, beyond recognizing visual semantics alone.
Existing methods introduce spatial structure into VLA models through several complementary directions.
Spatial Forcing \cite{li2025spatial} aligns implicit spatial representations during policy learning, while SpatialVLA \cite{qu2025spatialvlaexploringspatialrepresentations} incorporates spatially structured representations to improve action prediction.
Other approaches use explicit 3D information: SUGAR \cite{chen2024sugar} pretrains transferable 3D representations for robotic manipulation, Lift3D \cite{jia2024lift3d} lifts large-scale 2D representations into 3D policy features, and PointVLA \cite{li2026pointvla} injects point-based scene information into VLA models.
3D-VLA \cite{zhen20243dvla3dvisionlanguageactiongenerative} further combines 3D scene understanding with generative world modeling.

Feed-forward geometric foundation models such as VGGT and VGGT-$\Omega$
\cite{wang2025vggt,wang2026vggt}
provide a way to obtain geometry-aware representations by jointly aggregating multiple views and modeling their cross-view relationships. 
3D-Mix \cite{yu20263d} adaptively fuses VGGT-derived geometry with VLM features, while a recent study systematically examines how geometric foundation-model representations influence VLA action prediction \cite{yang2026understanding}.
These studies demonstrate the value of VGGT-derived geometry for robot control.
However, they primarily use geometric representations as perceptual inputs or spatial priors, rather than modeling their temporal evolution through a world-model objective.
GWM-VLA instead uses jointly aggregated VGGT-$\Omega$ representations as predictive states for action-conditioned latent world modeling.

\subsection{World Modeling for VLA}

World models provide VLA policies with predictive supervision by modeling future observations or representations. 
This encourages the policy to capture state changes caused by robot actions rather than relying only on the current observation.
World-action models (WAMs) integrate visual prediction and robot control within a unified model.
DreamZero \cite{ye2026world}, for example, jointly generates future video and robot actions as a closed-loop policy.
A recent comparative study reports strong robustness for several WAMs under distribution shifts \cite{zhang2026world}.
World models can also be used as external components for VLA policies.
For example, $\pi_{0.7}$ \cite{pi07} uses a separate generative world model to produce multi-view subgoal images, which are then provided to the VLA as additional context.

Explicit world models generate future images or videos, requiring them
to model both task-relevant changes and fine-grained appearance.
Latent world models instead predict future representations, providing
a more compact objective for policy learning.
VLA-JEPA \cite{sun2026vla} follows this approach by predicting future
visual latents conditioned on learned latent actions.
However, it encodes each camera view independently and combines their features through concatenation, leaving cross-view geometric relationships implicit.
It also introduces additional embodied-action representations for action generation rather than directly reusing the latent actions that condition the world model.
GWM-VLA instead models next-step transitions over jointly aggregated multi-view geometric states and uses the same latent-action representations for both latent prediction and continuous action generation.

\subsection{Latent Action Learning}

Latent action learning enables policy pretraining from videos without action annotations by representing visual transitions with implicit action variables \cite{chen2025moto,bu2025agibot}. 
LAPA \cite{ye2025latent} learns discrete latent actions for VLA pretraining, while UniVLA \cite{bu2025univla} introduces language-guided, task-centric latent actions. 
Villa-X \cite{chen2025villa} further grounds latent actions in physical dynamics by predicting proprioceptive states and jointly modeling latent and robot actions. 
However, transition-based objectives may capture dominant visual changes, such as camera motion and background variation, rather than only changes related to robot control \cite{zhang2026latent,lin2026pixels,nikulin2025latent}. 
MVP-LAM \cite{lee2026mvp} reduces viewpoint dependence through cross-view reconstruction. 
These methods primarily focus on action representation learning rather than modeling the future evolution of latent environment states.

\begin{figure*}[!t]
\centering
\includegraphics[width=1\linewidth]{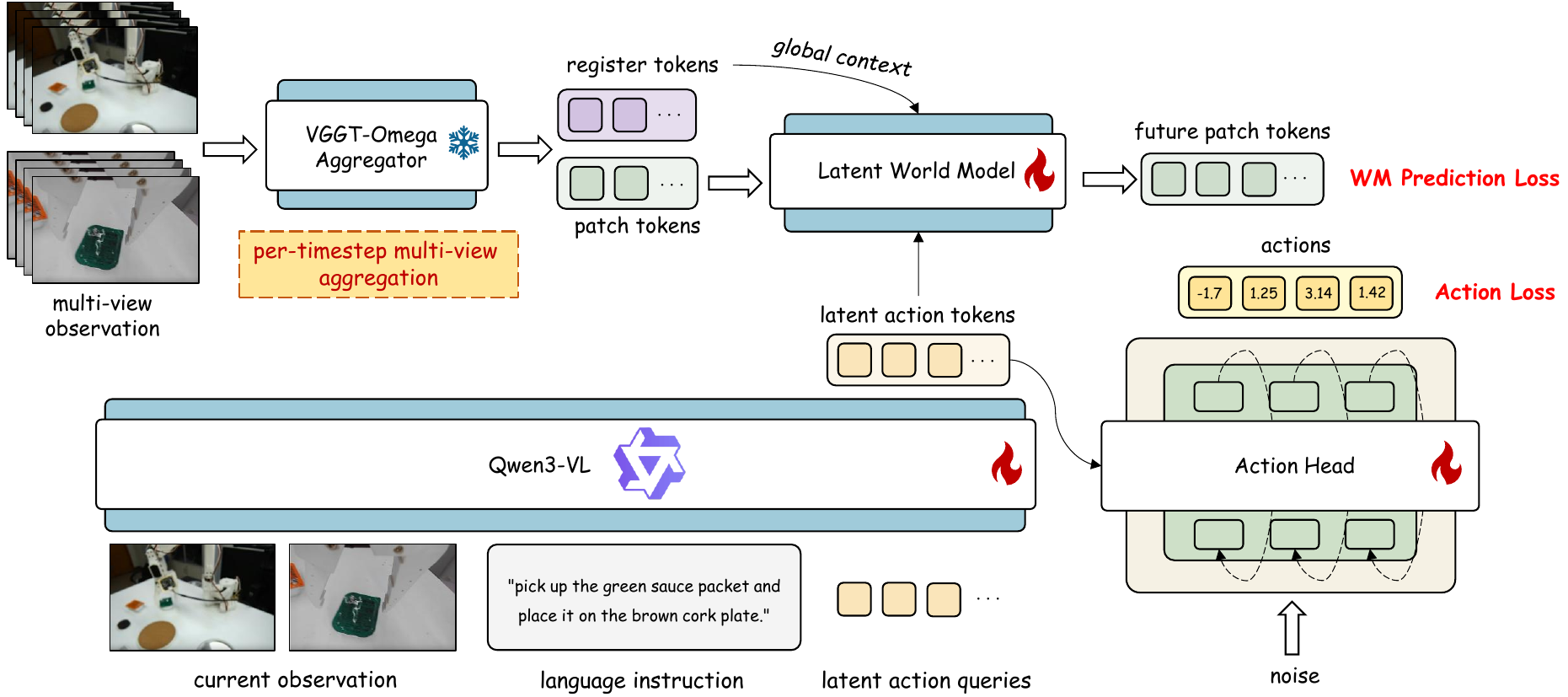}
\caption{Overall architecture of GWM-VLA. A frozen VGGT-$\Omega$ encoder aggregates multi-view observations at each timestep. Shared latent-action tokens condition both next-step latent prediction and flow-matching action generation. Snowflakes denote frozen modules, and flames denote trainable modules.}
\label{fig:architecture}
\end{figure*}

\section{Methodology}
\label{sec:method}

\subsection{Overview}
\label{sec:method_overview}

We consider a robot manipulation trajectory segment of horizon $T$ consisting of multi-view observations, a language instruction, proprioceptive states, and continuous robot actions:
\begin{equation}
    \boldsymbol{\tau}
    =
    \left(
    \mathbf{O}_{0:T},
    \ell,
    \mathbf{s}_{0:T},
    \mathbf{u}_{0:T-1}
    \right)
\end{equation}
where $T$ denotes the shared prediction and action horizon, and $\mathbf{O}_{t}$ is the multi-view observation:
\begin{equation}
    \mathbf{O}_{t}
    =
    \left\{
    I_{t}^{v}
    \right\}_{v=1}^{V}
\end{equation}
Here $\ell$ denotes the language instruction, $\mathbf{s}_{t}$ is the robot proprioceptive state, and $\mathbf{u}_{t}$ is the executed action.

As illustrated in Fig.~\ref{fig:architecture}, \textbf{GWM-VLA} consists of three main components:
(1) a geometry-aware multi-view encoder based on VGGT-$\Omega$,
(2) a latent world model with global context-conditioned target-view prediction, and
(3) a flow-matching action policy conditioned on a unified latent action representation.

GWM-VLA jointly encodes multi-view observations at each timestep with VGGT-$\Omega$, while processing different timesteps independently.
The latent world model performs causal, teacher-forced next-step prediction over the geometry-aware latent sequence, while the VLM-derived latent-action representation conditions both the latent world model and the flow-matching policy.

\subsection{Geometry-Aware Multi-View State Encoding}
\label{sec:geometry_encoder}

Generic visual encoders are not explicitly optimized to model the geometric relationships shared across multi-view observations. In contrast, robotic manipulation requires spatial reasoning about object geometry, camera geometry, and gripper-object interactions. We therefore employ a pretrained VGGT-$\Omega$ encoder to construct geometry-aware multi-view states. The VGGT-$\Omega$ aggregator is frozen during training.

At each timestep $t$, multi-view observations are jointly processed as a single geometric sample:
\begin{equation}
    \left\{
    \mathbf{R}_{t},
    \mathbf{P}_{t}
    \right\}
    =
    E_{\Omega}
    \left(
    \left\{
    I_{t}^{v}
    \right\}_{v=1}^{V}
    \right),
    \label{eq:vggt_encoding}
\end{equation}
where $E_{\Omega}$ denotes the VGGT-$\Omega$ encoder, and $\mathbf{R}_{t}$ and $\mathbf{P}_{t}$ denote the register and patch tokens, respectively.
Equation~\ref{eq:vggt_encoding} is applied independently at each timestep, while all views at the same timestep are jointly aggregated by the geometry encoder.

\subsection{Global Context-Conditioned Target-View Prediction}
\label{sec:world_model}

We perform next-step prediction in the geometry-aware latent space of VGGT-$\Omega$ rather than reconstructing future RGB observations.
Predicting the complete multi-view state would distribute the prediction objective across all camera streams.
We therefore apply the prediction loss to a selected target view and condition the latent world model on the corresponding target-view register tokens. Since both the patch and register tokens are extracted after joint multi-view aggregation, the model predicts only one view while still leveraging geometric information from multi-view observations.

Let $v^{\star}$ denote a target view selected according to the sensor configuration and task distribution.
After multi-view aggregation, we extract the corresponding register and patch tokens:
\begin{equation}
    \mathbf{r}_{t}
    =
    \mathbf{R}_{t}^{v^{\star}},
    \qquad
    \mathbf{p}_{t}
    =
    \mathbf{P}_{t}^{v^{\star}}.
    \label{eq:target_view_tokens}
\end{equation}

Here, $\mathbf{p}_{t}$ defines the target-view latent state to be predicted, while $\mathbf{r}_{t}$ provides global geometric context to the latent world model.
Both token sets correspond to the selected view but are produced after cross-view interaction within VGGT-$\Omega$.
In our experiments, we select the wrist view because its observations are closely coupled with end-effector motion and local gripper-object interactions.
We further condition the latent world model on a latent-action sequence $\mathbf{A}_{0:T-1}$ produced by the vision-language backbone and shared with the action policy.
Given the target-view patch tokens, target-view register tokens, and latent action tokens, the latent world model estimates the next target-view patch tokens at each timestep:
\begin{equation}
    \hat{\mathbf{p}}_{1:T}
    =
    F_{\phi}
    \left(
    \mathbf{p}_{0:T-1},
    \mathbf{r}_{0:T-1},
    \mathbf{A}_{0:T-1}
    \right),
    \label{eq:world_model}
\end{equation}
where $F_{\phi}$ denotes the action-conditioned next-step latent world model.

The latent world model employs time-causal attention. Tokens within the same timestep can interact bidirectionally, while tokens at timestep $t$ can attend only to tokens from timesteps no later than $t$. The attention mask is defined as
\begin{equation}
    M_{ij}
    =
    \begin{cases}
        0, & \tau(j) \leq \tau(i), \\
        -\infty, & \tau(j) > \tau(i),
    \end{cases}
    \label{eq:causal_mask}
\end{equation}
where $\tau(i)$ denotes the timestep associated with token $i$.

During training, we adopt teacher forcing. The ground-truth latent state $\mathbf{p}_{t}$ is provided when predicting the next state $\mathbf{p}_{t+1}$. The supervision targets are obtained by encoding future multi-view observations with the same per-timestep geometry encoder:
\begin{equation}
    \mathbf{p}_{t+1}
    =
    E_{\Omega}^{P,v^{\star}}
    \left(
    \mathbf{O}_{t+1}
    \right),
\end{equation}
where $E_{\Omega}^{P,v^{\star}}$ denotes the target-view patch-token output of VGGT-$\Omega$.
Together with the causal attention mask in the latent world model, this yields an action-conditioned prediction objective over the geometry-aware states defined above.

We optimize the world model using an $\ell_{1}$ latent prediction objective:
\begin{equation}
    \mathcal{L}_{\mathrm{wm}}
    =
    \frac{1}{T}
    \sum_{t=0}^{T-1}
    \left\|
    \hat{\mathbf{p}}_{t+1}
    -
    \mathbf{p}_{t+1}
    \right\|_{1},
    \label{eq:wm_loss}
\end{equation}

\subsection{Unified Latent Action Representation}
\label{sec:latent_action}

To connect latent dynamics learning with robot control, GWM-VLA uses a shared latent-action representation to condition both the world model and the action policy.
We adopt Qwen3-VL-2B \cite{bai2025qwen3vltechnicalreport} as the vision-language backbone. Given the current multi-view observation and the language instruction, we insert timestep-grouped learnable latent action query tokens:
\begin{equation}
\begin{aligned}
    \mathcal{Q}^{A}
    &=
    \left[
    \mathcal{Q}_{0}^{A},
    \mathcal{Q}_{1}^{A},
    \ldots,
    \mathcal{Q}_{T-1}^{A}
    \right], \\
    \mathcal{Q}_{t}^{A}
    &=
    \left[
    q_{t,1}^{A},
    \ldots,
    q_{t,K}^{A}
    \right].
\end{aligned}
\end{equation}
All timestep-grouped latent-action representations are produced from the current observation $\mathbf{O}_{0}$; $t$ indexes positions in the future action horizon rather than separately observed policy inputs.
The horizon $T$ is shared by latent prediction and action generation, providing one timestep-indexed latent-action group for each predicted transition and action-chunk position.

The corresponding latent action representations are obtained from the hidden states of the vision-language model:
\begin{equation}
    \mathbf{A}_{0:T-1}
    =
    Q_{\theta}
    \left(
    \mathbf{O}_{0},
    \ell,
    \mathcal{Q}^{A}
    \right),
    \label{eq:latent_action}
\end{equation}
where
\begin{equation}
    \mathbf{A}_{t}
    =
    \left[
    \mathbf{a}_{t,1},
    \ldots,
    \mathbf{a}_{t,K}
    \right]
\end{equation}
contains $K$ latent action tokens associated with timestep $t$.

Unlike VLA-JEPA, which introduces additional embodied action queries for action generation, GWM-VLA directly reuses the latent-action sequence $\mathbf{A}_{0:T-1}$ to condition both the latent world model and the flow-matching action head. 
This shared design allows the latent prediction and robot-action objectives to jointly shape the same latent-action representations, encouraging them to capture dynamics relevant to robot control.

\subsection{Conditional Flow-Matching Action Head}
\label{sec:action_policy}

To capture the multimodal distribution of continuous robot actions, we employ a conditional flow-matching head \cite{lipman2023flowmatchinggenerativemodeling} to generate action chunks. 
The action head is conditioned on the shared latent-action sequence $\mathbf{A}_{0:T-1}$ and current proprioceptive state $\mathbf{s}_{0}$, and generates an action chunk over horizon $T$.
Given a ground-truth action chunk $\mathbf{u}_{0:T-1}$, Gaussian noise $\boldsymbol{\epsilon} \sim \mathcal{N}(\mathbf{0}, \mathbf{I})$, flow time $\gamma \sim \mathcal{U}(0,1)$, and $\mathbf{u}_{\gamma}=(1-\gamma)\boldsymbol{\epsilon}+\gamma\mathbf{u}_{0:T-1}$, we optimize
\begin{equation}
\begin{aligned}
    \mathcal{L}_{\mathrm{action}}
    =
    \mathbb{E}
    \Big[
    \big\|&
    v_{\psi}
    \left(
    \mathbf{u}_{\gamma},
    \gamma
    \mid
    \mathbf{A}_{0:T-1},
    \mathbf{s}_{0}
    \right)
    \\
    &-
    \left(
    \mathbf{u}_{0:T-1}
    -
    \boldsymbol{\epsilon}
    \right)
    \big\|_{2}^{2}
    \Big].
\end{aligned}
\label{eq:action_loss}
\end{equation}

Full training and inference details are provided in the appendix.

\subsection{Joint World-Model and Policy Learning}
\label{sec:joint_objective}

GWM-VLA jointly optimizes teacher-forced next-step latent prediction and robot action generation:
\begin{equation}
    \mathcal{L}
    =
    \mathcal{L}_{\mathrm{action}}
    +
    \lambda
    \mathcal{L}_{\mathrm{wm}},
    \label{eq:total_loss}
\end{equation}
where $\lambda$ balances the predictive and control objectives.
\begin{table*}[t]
    \centering
    \setlength{\tabcolsep}{5pt}
    \begin{tabular}{@{}lccccc@{}}
        \toprule
        Method
        & Spatial
        & Object
        & Goal
        & LIBERO-10
        & Avg. \\
        \midrule
        LAPA~\cite{ye2025latent}
        & 73.8 & 74.6 & 58.8 & 55.4 & 65.7 \\
        UniVLA~\cite{bu2025univla}
        & 96.5 & 96.8 & 95.6 & 92.0 & 95.2 \\
        OpenVLA-OFT~\cite{kim2025fine}
        & \underline{97.6}
        & 98.4
        & \underline{97.9}
        & \textbf{94.5}
        & \textbf{97.1} \\
        $\pi_0$~\cite{black2024pi_0}
        & 96.8
        & 98.8
        & 95.8
        & 85.2
        & 94.2 \\
        $\pi_0$-FAST~\cite{pertsch2025fast}
        & 96.4 & 96.8 & 88.6 & 60.2 & 85.5 \\
        CoT-VLA~\cite{zhao2025cot}
        & 87.5 & 91.6 & 87.6 & 69.0 & 81.1 \\
        WorldVLA~\cite{cen2025worldvla}
        & 87.6 & 96.2 & 83.4 & 60.0 & 81.8 \\
        villa-X~\cite{chen2025villa}
        & 97.5 & 97.0 & 91.5 & 74.5 & 90.1 \\
        GR00T N1~\cite{bjorck2025gr00t}
        & 94.4 & 97.6 & 93.0 & 90.6 & 93.9 \\
        $\pi_{0.5}$~\cite{intelligence2025pi_}
        & \textbf{98.8}
        & 98.2
        & \textbf{98.0}
        & 92.4
        & \underline{96.9} \\
        VLA-JEPA~\cite{sun2026vla} w/o human videos
        & 94.8
        & \textbf{99.6}
        & 95.8
        & 94.0
        & 96.1 \\
        \midrule
        \ours{}
        & 96.8
        & \underline{99.0}
        & \textbf{98.0}
        & \underline{94.4}
        & \textbf{97.1} \\
        \bottomrule
    \end{tabular}
    \caption{LIBERO success rates (\%). Bold and underlined values denote the best and second-best results, respectively.}
    \label{tab:libero_main}
\end{table*}

\begin{table*}[t]
    \centering
    \setlength{\tabcolsep}{3.5pt}
    \begin{tabular}{@{}lcccccccc@{}}
        \toprule
        Method
        & Camera
        & Robot
        & Language
        & Light
        & Background
        & Noise
        & Layout
        & Avg. \\
        \midrule
        UniVLA~\cite{bu2025univla}
        & 1.8
        & 46.2
        & 69.6
        & 69.0
        & 81.0
        & 21.2
        & 31.9
        & 42.9 \\

        OpenVLA-OFT~\cite{kim2025fine}
        & 56.4
        & 31.9
        & \underline{79.5}
        & \underline{88.7}
        & \textbf{93.3}
        & \underline{75.8}
        & 74.2
        & \underline{69.6} \\

        $\pi_0$~\cite{black2024pi_0}
        & 13.8
        & 6.0
        & 58.8
        & 85.0
        & 81.4
        & \textbf{79.0}
        & 68.9
        & 53.6 \\

        $\pi_0$-FAST~\cite{pertsch2025fast}
        & \textbf{65.1}
        & 21.6
        & 61.0
        & 73.2
        & 73.2
        & 74.4
        & 68.8
        & 61.6 \\

        WorldVLA~\cite{cen2025worldvla}
        & 0.1
        & 27.9
        & 41.6
        & 43.7
        & 17.1
        & 10.9
        & 38.0
        & 25.0 \\

        VLA-JEPA~\cite{sun2026vla} w/o human videos
        & 40.3
        & \textbf{55.7}
        & 72.9
        & 88.2
        & 70.5
        & 38.2
        & \underline{74.6}
        & 62.9 \\
        \midrule

        \ours{}
        & \underline{57.9}
        & \underline{54.7}
        & \textbf{89.8}
        & \textbf{95.4}
        & \underline{90.8}
        & 72.5
        & \textbf{77.1}
        & \textbf{76.9} \\
        \bottomrule
    \end{tabular}
    \caption{LIBERO-Plus success rates (\%). Bold and underlined values denote the best and second-best results, respectively.}
    \label{tab:libero_plus_main}
\end{table*}

We use $\lambda=0.1$ as the default world-model loss weight in the main experiments; a sensitivity analysis is provided in the appendix.
This joint objective encourages the latent action to encode action-conditioned interaction changes while remaining grounded in executable robot controls.
The world-model loss serves as an auxiliary training objective and is not used during deployment.

\section{Experiments}
\label{sec:exper}

We evaluate \ours{} in simulation and on a real robot, including robustness benchmarks, controlled ablations, and qualitative probing of predicted latent states.

\subsection{Experimental Setup}
\label{sec:experimental_setup}

\paragraph{Implementation details.}
\ours{} uses Qwen3-VL-2B as the vision-language backbone and VGGT-$\Omega$ as the geometry-aware multi-view encoder.
The VGGT-$\Omega$ aggregator is frozen throughout training, while Qwen3-VL, the latent world model, and the action head are fully trainable.
Unless otherwise specified, we use the final VGGT-$\Omega$ representation, set the world-model loss weight to $\lambda=0.1$, and directly condition the action head on the shared latent-action tokens.
Pretraining and simulation fine-tuning are conducted using $8$ NVIDIA A800 GPUs, while ablation experiments are conducted on a single NVIDIA RTX 6000D GPU.
Additional training details and hyperparameter analyses are provided in the appendix.

\paragraph{Datasets.}
We pretrain \ours{} on DROID~\cite{khazatsky2024droid}, a large-scale action-labeled robot dataset containing approximately $76{,}000$ demonstration trajectories collected across multiple institutions and operators.
The dataset covers diverse manipulation tasks, camera viewpoints, workspace layouts, and visual backgrounds, providing broad variation for learning transferable robot representations.
For simulation experiments, we fine-tune the pretrained model on the standard demonstration sets provided by LIBERO.

\paragraph{Benchmarks.}
We conduct simulation experiments on the LIBERO~\cite{liu2023libero} and LIBERO-Plus~\cite{fei2025libero} benchmarks.
LIBERO uses the Franka Emika Panda arm and includes four task suites, \textsc{Spatial}, \textsc{Object}, \textsc{Goal}, and \textsc{LIBERO-10}, which evaluate spatial reasoning, object-centric manipulation, goal-conditioned control, and long-horizon execution.
We use it as the in-distribution simulation benchmark.
LIBERO-Plus extends the same task families with seven perturbation dimensions, including camera viewpoint, robot initial state, language instruction, illumination, background, visual noise, and object layout, and serves as our out-of-distribution robustness benchmark.
For both benchmarks, we report task success rate; LIBERO results are averaged over suites, while LIBERO-Plus results are averaged over perturbation dimensions.

\paragraph{Baselines.}
We compare \ours{} with representative VLA, latent-action, and world-model-based methods, including LAPA~\cite{ye2025latent}, UniVLA~\cite{bu2025univla}, OpenVLA-OFT~\cite{kim2025fine}, $\pi_0$~\cite{black2024pi_0}, $\pi_0$-FAST~\cite{pertsch2025fast}, CoT-VLA~\cite{zhao2025cot}, WorldVLA~\cite{cen2025worldvla}, villa-X~\cite{chen2025villa}, GR00T N1~\cite{bjorck2025gr00t}, $\pi_{0.5}$~\cite{intelligence2025pi_}, and VLA-JEPA~\cite{sun2026vla}.
Because \ours{} is pretrained only on robot demonstrations, robot-only VLA-JEPA without human-video pretraining serves as the most directly comparable baseline.

\subsection{Simulation Benchmark Experiments}
\label{sec:simulation_evaluation}

\paragraph{LIBERO.}
\label{sec:libero_results}

As shown in Table~\ref{tab:libero_main}, \ours{} achieves a
$97.1\%$ average success rate, tying OpenVLA-OFT for the best overall
performance and outperforming the robot-only VLA-JEPA baseline by
$1.0$ percentage point.
Its consistently strong performance across all four task suites shows
that geometry-aware latent world modeling preserves effective
in-distribution control while introducing additional geometric and
predictive supervision.

\paragraph{LIBERO-Plus.}
\label{sec:libero_plus_results}

Table~\ref{tab:libero_plus_main} reports the results on LIBERO-Plus. 
\ours{} achieves a state-of-the-art average success rate of $76.9\%$, outperforming OpenVLA-OFT by $7.3$ percentage points and the robot-only VLA-JEPA baseline by $14.0$ percentage points.
It ranks first under language, lighting, and object-layout perturbations,
and second under camera-viewpoint, robot-initial-state, and background
perturbations.
The largest gains over VLA-JEPA occur under visual noise, background,
camera-viewpoint, and language changes.
This broad improvement is consistent with jointly aggregated geometric
states providing more stable spatial representations under visual and
environmental shifts.

\subsection{Real-World Experiments}
\label{sec:real_world_experiments}

\paragraph{Setup.}
We further evaluate \ours{} on the SO-101 robot using the open-source LeRobot deployment stack \cite{cadenelerobot}.
We collect $100$ teleoperated robot demonstrations over five training pick-and-place tasks, and all methods are adapted using this same demonstration set.
Each evaluation task is attempted $10$ times.
We consider three settings: in-distribution (ID), consisting of three seen instructions from the training task set under nominal layouts; OOD task, consisting of three held-out object--receptacle recombinations; and OOD object layouts, consisting of the same three ID instructions evaluated under novel object arrangements.
We compare against $\pi_0$~\cite{black2024pi_0} and $\pi_{0.5}$~\cite{intelligence2025pi_}.

\begin{figure}[t]
    \centering
    \includegraphics[width=\columnwidth]{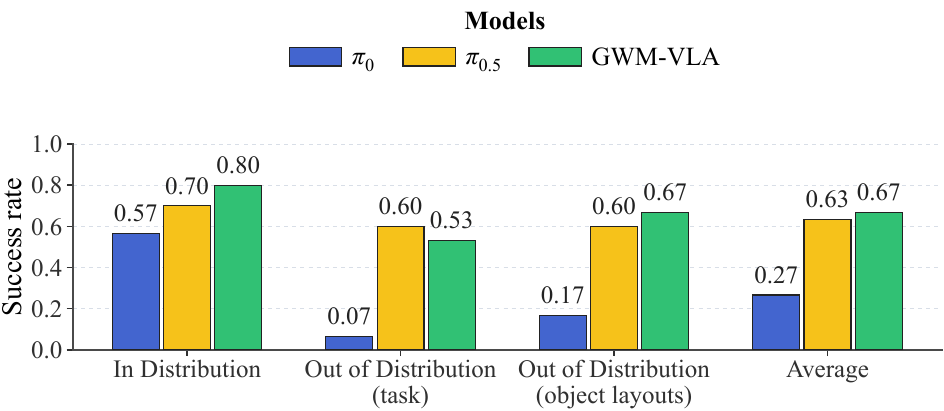}
    \caption{Real world experimental results under ID, OOD-task, and OOD-layout settings.}
    \label{fig:real_world_main}
\end{figure}

\paragraph{Results.}
Figure~\ref{fig:real_world_main} summarizes the real-world results.
\ours{} obtains the highest observed average success rate in our real-world evaluation and is most effective in the ID and OOD-layout settings.
These settings stress execution accuracy and spatial robustness under limited real-world demonstrations, and the results are consistent with geometry-aware state prediction providing a useful inductive bias.
The OOD-task setting presents a different challenge: the policy must recombine object and receptacle concepts into held-out instructions.
Here $\pi_{0.5}$ performs best at $60.0\%$, while \ours{} reaches $53.3\%$ and substantially outperforms $\pi_0$.
This result suggests that spatial robustness and compositional instruction generalization remain distinct challenges.

\subsection{Ablation Studies}
\label{sec:ablations}

We conduct ablations on LIBERO (spatial suite) to investigate the key design choices of \ours{}.
The default configuration uses the final VGGT-$\Omega$ layer and direct latent-action conditioning.
These ablations are lightweight diagnostic runs trained only on LIBERO (spatial suite) using a single NVIDIA RTX 6000D GPU, batch size $16$, and $10{,}000$ optimization steps.
They are therefore intended to compare design variants under the same restricted setting, rather than to be directly compared with the main pretraining-plus-fine-tuning results.

\subsubsection{Effect of Target-View Selection}

\begin{figure}[!t]
    \centering
    \includegraphics[width=0.90\columnwidth]{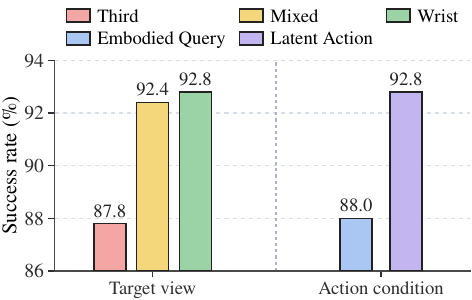}
    \caption{Diagnostic ablations on LIBERO (spatial suite) for target-view selection and action conditioning.}
    \label{fig:ablation_main}
\end{figure}

We study how the choice of target view affects policy performance in the LIBERO-Spatial setting.
As shown in Figure~\ref{fig:ablation_main} (left), wrist-view prediction achieves a $92.8\%$ success rate, outperforming third-person-view prediction at $87.8\%$.
Mixed-view prediction, which samples $80\%$ wrist-view targets and $20\%$ third-person-view targets, achieves a comparable but slightly lower success rate of $92.4\%$.
These results suggest that the wrist view is the most effective target among the evaluated configurations.
We attribute the advantage of the wrist view to its close relation to end-effector motion and local gripper-object interactions.
Because its tokens are obtained after joint multi-view aggregation, the prediction target retains global geometric context while emphasizing local manipulation dynamics.

\subsubsection{Effect of the Representation Depth}
\begin{figure}[t]
    \centering
    \includegraphics[width=0.90\columnwidth]{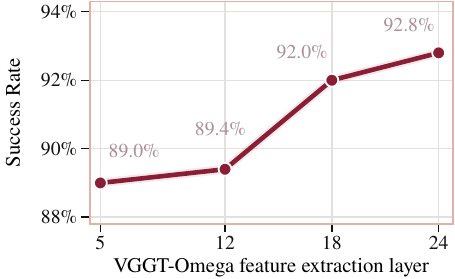}
    \caption{Effect of VGGT-$\Omega$ representation depth on LIBERO (spatial suite).}
    \label{fig:ablation_layer}
\end{figure}

The representation depth here refers to the VGGT-$\Omega$ layer from which the patch and register tokens are extracted to construct the latent state. 
Prior studies show that VGGT exhibits strong layer-wise differences: cross-view correspondences and epipolar geometry emerge in the middle layers and are further refined toward later layers \cite{bratulic2026geometric,you2026regimevggt}. 
We therefore compare VGGT-$\Omega$ representations extracted from Layers $5$, $12$, $18$, and $24$.
As shown in Figure~\ref{fig:ablation_layer}, success rate increases with the layer used to construct the latent state, and Layer $24$ performs best. This result suggests that latent world modeling benefits from the more refined geometric representations available in later VGGT-$\Omega$ layers. We therefore use Layer $24$ by default.

\subsubsection{Effect of the Shared Latent Action Representation}

We compare directly conditioning the action head on latent action tokens with introducing an additional embodied-action query.
The action-conditioning results are shown in Figure~\ref{fig:ablation_main} (right).
Under the controlled LIBERO (spatial suite) setting, direct latent-action conditioning outperforms the embodied-query variant by 4.8 percentage points. This result supports directly sharing the latent-action tokens between latent prediction and action generation.

\subsection{Qualitative Probing of Predicted Latent States}
\label{sec:world_model_visualization}

\begin{figure}[!t]
    \centering
    \includegraphics[width=\columnwidth]{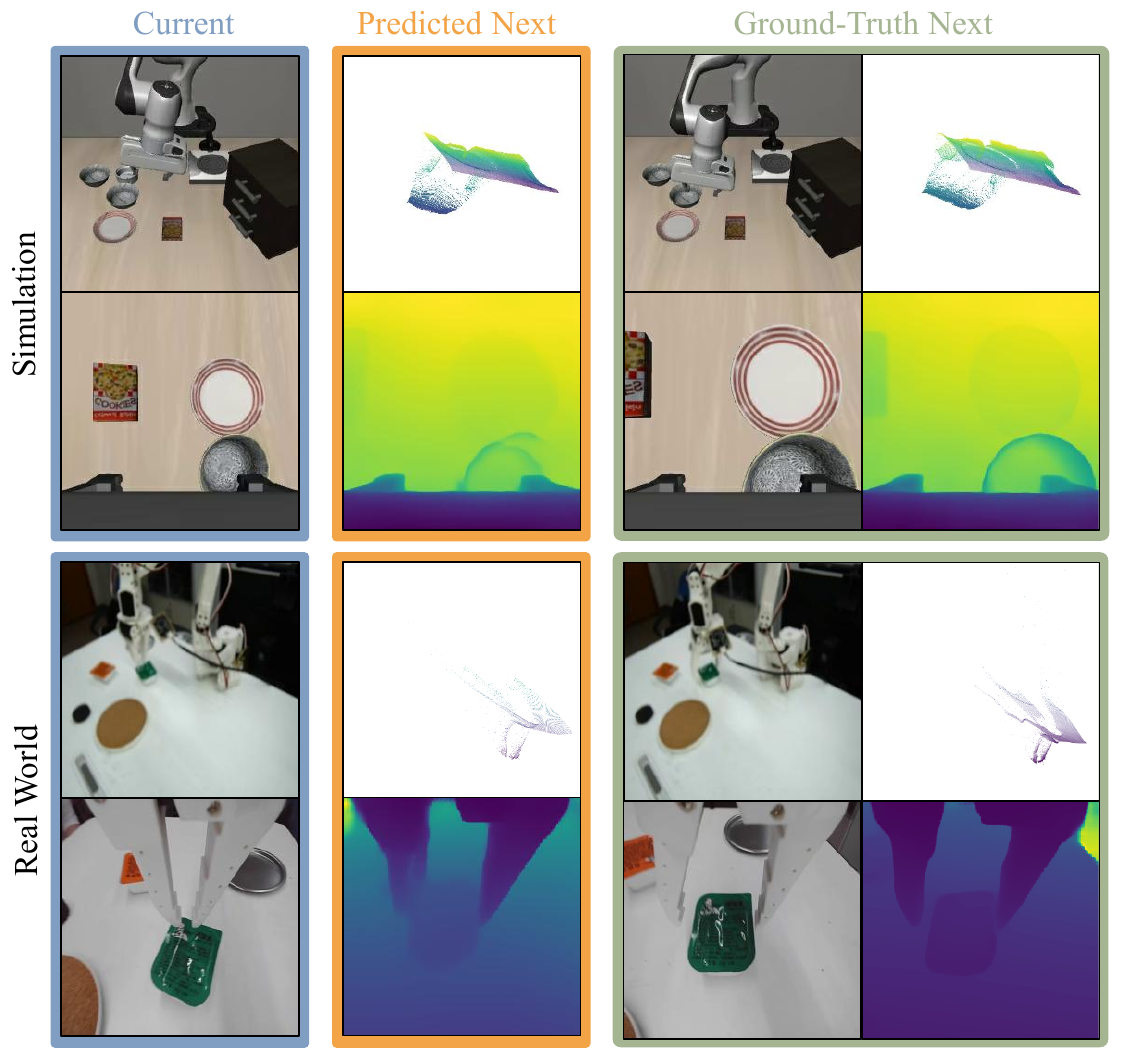}
    \caption{Qualitative probing of predicted next-step latent states using a separately trained and frozen depth decoder and camera-coordinate point clouds.}
    \label{fig:wm_visualization}
\end{figure}

Figure~\ref{fig:wm_visualization} qualitatively probes whether predicted latent states retain next-step geometric information.
A visualization decoder is trained on observed next-step VGGT-$\Omega$ tokens to match depth maps from the frozen dense head, then frozen and applied to predicted tokens.
The decoded results preserve the main scene layout and geometry around the gripper and manipulated objects, although fine details are smoothed.
These results are qualitative probes rather than measurements of 3D reconstruction accuracy.

\FloatBarrier

\section{Conclusion and Limitations }
\label{sec:conclusion}

We propose GWM-VLA, a geometry-aware latent world modeling framework for Vision-Language-Action learning.
GWM-VLA jointly aggregates multi-view observations at each timestep with VGGT-$\Omega$ to construct geometry-aware states, predicts next-step target-view patch tokens using the corresponding register tokens as global geometric context, and uses shared latent-action representations for both latent prediction and continuous action generation.
Across LIBERO and LIBERO-Plus, GWM-VLA achieves strong in-distribution performance and substantial robustness improvements under visual and environmental shifts.
Real-world experiments further demonstrate its effectiveness under limited demonstrations and novel object layouts.
Despite these results, GWM-VLA relies on multi-view observations at the same timestep to construct geometry-aware latent states, which limits its direct applicability to large-scale human-video pretraining, where such multi-view observations are often unavailable.
It also uses a fixed target view during training.
Although the wrist view performs best on LIBERO-Spatial, the optimal target may vary across tasks and sensor configurations.
Future work will extend geometry-aware latent world modeling to single-view video data, and explore adaptive target-view selection and multi-target prediction.


\bibliography{aaai2027}


\end{document}